\documentclass[10pt,twocolumn,letterpaper]{article}

\usepackage{cvpr}
\usepackage{times}
\usepackage{epsfig}
\usepackage{graphicx}
\usepackage{amsmath}
\usepackage{amssymb}

\usepackage{algorithm,algorithmic}
\usepackage{paralist, tabularx}

\newcommand{\ignore}[1]{}

\cvprfinalcopy % *** Uncomment this line for the final submission

\def\cvprPaperID{****} % *** Enter the CVPR Paper ID here

\begin{document}

%%%%%%%%% TITLE
\title{Diversified Texture Synthesis with Feed-forward Networks}

\author{Yijun Li$^1$, Chen Fang$^2$, Jimei Yang$^2$, Zhaowen Wang$^2$, Xin Lu$^2$, and Ming-Hsuan Yang$^1$ \\ \\$^1$University of California, Merced~~~~~~~~~~~~~~~~$^2$Adobe Research}

%\author{First Author\\
%Institution1\\
%Institution1 address\\
%{\tt\small firstauthor@i1.org}
% For a paper whose authors are all at the same institution,
% omit the following lines up until the closing ``}''.
% Additional authors and addresses can be added with ``\and'',
% just like the second author.
% To save space, use either the email address or home page, not both
%\and
%Second Author\\
%Institution2\\
%First line of institution2 address\\
%{\tt\small secondauthor@i2.org}
%}

\maketitle
%\thispagestyle{empty}

%%%%%%%%% ABSTRACT
\begin{abstract}

Recent progresses on deep discriminative and generative modeling have shown promising results on texture synthesis.
However, existing feed-forward based methods trade off generality for efficiency, which suffer from many issues, such as shortage of generality (i.e., build one network per texture), lack of diversity (i.e., always produce visually identical output) and suboptimality (i.e., generate less satisfying visual effects).
In this work, we focus on solving these issues for improved texture synthesis. % and style transfer.
We propose a deep generative feed-forward network which enables efficient synthesis of multiple textures within one single network
and meaningful interpolation between them.
%
%Given $M$ textures, the resulting network is much lighter than $M$ existing single-texture networks but generates textures of comparable quality.
%
Meanwhile, a suite of important techniques are introduced to achieve better convergence and diversity.
With extensive experiments, we demonstrate the effectiveness of the proposed model and techniques for synthesizing a large number of textures and show its applications with the stylization.

\end{abstract}

%%%%%%%%% BODY TEXT
\section{Introduction}
\begin{figure*}[t]
\begin{center}
\includegraphics[width=17cm]{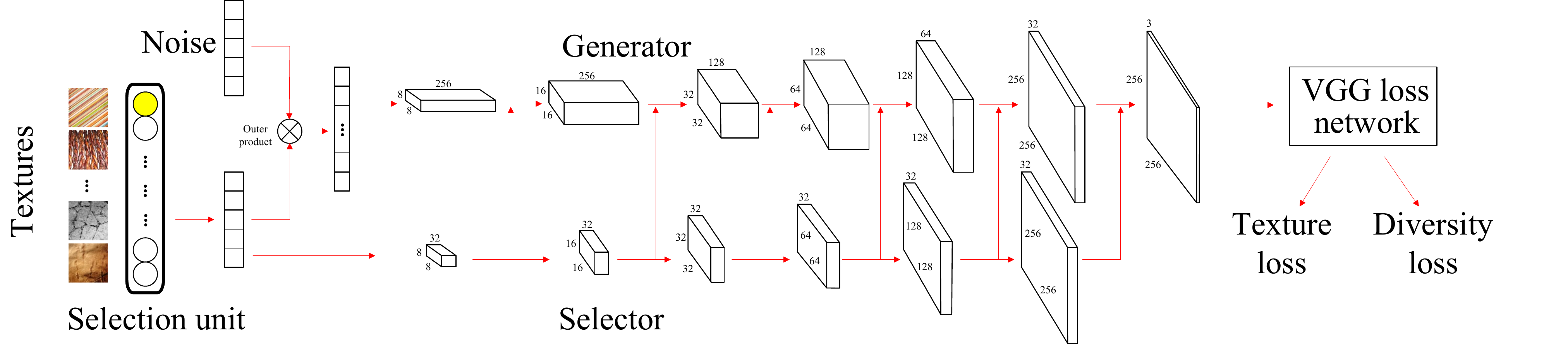}
\end{center}
\caption{Architecture of the proposed multi-texture synthesis network.
It consists of a generator and a selector network.
}
\label{fig:framework}
\end{figure*}
Impressive neural style transfer results by Gatys et al.~\cite{GatysTransfer-CVPR2016} have recently regained great interests from computer vision, graphics and machine learning communities for the classic problem of texture synthesis~\cite{Efros1-ICCV1999,Efros2-SIGGRAPH2001,wei-2001-texture}.
Considering the expensive optimization process in~\cite{GatysTransfer-CVPR2016}, a few attempts have been made to develop feed-forward networks to efficiently synthesize a texture image or a stylized image~\cite{Perceptual-ECCV2016,Texturenet-ICML2016}.
%
%These methods often synthesize one texture per network. This problem setup is inherited from Gatys et al.~\cite{GatysTransfer-CVPR2016} as they optimize one style in each time. But it is actually more computation- and memory-efficient to learn one network that is able to generate texture samples of many types.
However, these methods often suffer from many issues, including shortage of generality (i.e., build one network per texture), lack of diversity (i.e., always produce visually identical output) and suboptimality (i.e., generate less satisfying visual effects).

In this paper, we propose a deep generative network for synthesizing diverse outputs of multiple textures in a single network.
Our network architecture, inspired by~\cite{radford-2015-dcGAN,dosovitskiy-2015-learning}, takes a noise vector and a selection unit as input to generate texture images using up-convolutions.
The selection unit is a one-hot vector where each bit represents a texture type and provides users with a control signal to switch between different types of textures to synthesize.
More importantly, such a multi-texture synthesis network facilitates generating new textures by interpolating with the selection units.
Meanwhile, the noise vector is intended to drive the network to generate diverse samples even from a single exemplar texture image.

However, learning such a network is a challenging task.
First, different types of textures have quite different statistical characteristics, which are partially reflected by the varied magnitudes of texture losses (i.e., the Gram matrix-based losses introduced in~\cite{GatysTexture-NIPS2015,GatysTransfer-CVPR2016} to measure style similarity) across different feature layers.
Second, the convergence rates for fitting different textures are inherently different due to their drastic visual difference and semantic gaps. As a result, the overall difficulty of learning such a network is determined by the variation among the textures and the complexity of individual textures.
Third, the network often encounters the ``explain-away'' effect that the noise vector is marginalized out and thus fails to influence the network output.
Specifically, the network is not able to generate diverse samples of a given texture, and it often means overfitting to a particular instance.

In this work, we propose a suite of effective techniques to help the network generate diverse outputs of higher quality. %(Figure~\ref{fig:Gram_mean_subtraction},~\ref{fig:Diversity},~\ref{fig:Incremental60_single})
%
%Trained with this loss, our generator is capable of synthesizing different results with large variations for each texture. This will give users more freedom to select their favorite ones by sampling a noise vector.
We first improve the Gram matrix loss by subtracting the feature mean, so that the newly designed loss is more stable in scale and adds stability to the learning.
%\jimei{Please check if the sentence about Gram matrix is correct.}
%
Second, in order to empower the network with the ability of generating diverse samples and further prevent overfitting, we aim to correlate the output samples with the input noise vector. Specifically, we introduce a diversity loss that penalizes the feature similarities of different samples in a mini-batch.
Third, we show that a suitable training strategy is critical for the network to converge better and faster.
Thus we devise an incremental learning algorithm that expose new training textures sequentially to the learner network.
We start from learning to synthesize one texture and only incorporate next new unseen texture into the learning when the previous texture can be well generated.
As such, the network gradually learns to synthesize new textures while retaining the ability to generate all previously
seen textures.
%
%We show that these techniques can be also applied to existing single-texture networks~\cite{Perceptual-ECCV2016,Texturenet-ICML2016} for both synthesis and transfer for improved results.

The contributions of this work are threefold:

\begin{compactitem}
\item We propose a generative network to synthesize multiple textures in a user-controllable manner.
\item A diversity loss is introduced to prevent the network from being trapped in a degraded solution and more importantly it allows the network to generate diverse texture samples.
\item The incremental learning algorithm is demonstrated to be effective at training the network to synthesize results of better quality.
\end{compactitem}

\section{Related Work}

%In this section, we mainly review typical techniques of texture synthesis and also summarize works that apply synthesis to the style transfer.

\paragraph{Traditional synthesis models.} Texture synthesis methods are broadly categorized as non-parametric or parametric.
Parametric methods~\cite{heeger-1995-pyramid,portilla-2000-parametric} for texture synthesis
aim to represent textures through proper statistical models, with the assumption that two images can be visually similar
when certain image statistics match well~\cite{julesz-1962visual}.
%
%Since Julesz et. al.~\cite{julesz-1962visual} suggested that two textures will be perceived by observers to be the same if proper statistics of these images match.
%
The synthesis procedure starts from a random noise image and gradually coerces it to have the same relevant statistics as the given example.
The statistical measurement is either based on marginal filter response histograms~\cite{de-1997-multiresolution,heeger-1995-pyramid} at different scales or more complicated joint responses~\cite{portilla-2000-parametric}.
However, exploiting proper image statistics is challenging for parametric models especially when synthesizing structured textures.

Alternatively, non-parametric models~\cite{Efros1-ICCV1999,Efros2-SIGGRAPH2001,kwatra-2003-graphcut,wei-2000-fast}
focus on growing a new image from an initial seed and regard the given texture example as a source pool to constantly sample similar pixels or patches. This is also the basis of earlier texture transfer algorithms~\cite{Efros2-SIGGRAPH2001,lee-2010-directional,ashikhmin-2003-fast,Hertz-2001-analogy}.
Despite its simplicity, these approaches can be slow and subject to non-uniform pattern distribution.
More importantly, these methods aim at growing a perfect image instead of building rich models to understand textures.

\paragraph{Synthesis with neural nets.}~The success of deep CNNs in discriminative tasks~\cite{krizhevsky-2012-alexnet,russakovsky-2015-imagenet} has
attracted much attention for image generation.
Images can be reconstructed by inverting features~\cite{mahendran-CVPR2015-Inverting,Doso-CVPR2016-Inverting,Doso-NIPS2016-Generation}, synthesized by matching features, or even generated from noise~\cite{goodfellow-2014-GAN,radford-2015-dcGAN,denton-2015-LapGAN}.
Synthesis with neural nets is essentially a parametric approach, where intermediate network outputs provide rich and effective image statistics.
Gatys et al. \cite{GatysTexture-NIPS2015} propose that two textures are perceptually similar if their features extracted by a pre-trained CNN-based classifier
share similar statistics.
Based on this, a noise map is gradually optimized to a desired output that matches the texture example in the CNN feature space.

Subsequent methods~\cite{Perceptual-ECCV2016,Texturenet-ICML2016} accelerate this optimization procedure by formulating the generation as learning a feed-forward network.
These methods train a feed-forward network by minimizing the differences between statistics of the ground truth and the generated image. In particular, image statistics was measured by intermediate outputs of a pre-trained network.
Further improvements are made by other methods that follow either optimization based~\cite{MrfTransfer-CVPR2016,Frigo-2016-CVPR,Gatys-PreservingColor-2016} or feed-forward based~\cite{MGAN-ECCV2016,InstanceBN-arxiv-2016} framework.
However, these methods are limited by the unnecessary requirement of training one network per texture.
Our framework also belongs to the feed-forward category but
synthesizes diverse results for multiple textures in one single network.

% see if we need to cite this concurrent work
A concurrent related method recently proposed by Dumoulin et al.~\cite{GoogleMultiTexture-2016} handles
multi-style transfer in one  network by specializing scaling and shifting parameters after normalization to each specific texture.
Our work differs from~\cite{GoogleMultiTexture-2016} mainly in two aspects. First, we employ a different approach in representing textures. We represent textures as bits in a one-hot selection unit and as a continuous embedding vector within the network. Second, we propose diversity loss and incremental training scheme in order to achieve better convergence and output diverse results. Moreover, we demonstrate the effectiveness of our method on a much larger set of textures (e.g., 300)
whereas ~\cite{GoogleMultiTexture-2016} develops a network for 32 textures.

\section{Proposed Algorithm}

We show the network architecture of the proposed model in Figure~\ref{fig:framework}.
The texture synthesis network (bottom part) has two inputs, a noise vector and a selection unit, while the upper part in blue dash line boxes are modules added for extending our model to style transfer.
The noise vector is randomly sampled from a uniform distribution, and the selection unit is a one-hot vector, where each bit represents a texture in the given texture set.
The network consists of two streams: the generator and the selector. The generator is responsible for synthesis and the selector is guiding the generator towards the target texture, conditioned on the activated bit in the selection unit.

Given $M$ target textures, we first map the $M$ dimensional selection unit to a lower dimensional selection embedding.
Then we compute the outer product of the noise vector and selection embedding.
After the outer-product operation, we reshape the result as a bunch of $1\times1$ maps and then use the \emph{SpatialFullConvolution} layer to convolve them to a larger spatial representation with
numerous feature maps.
After a series of nearest-neighbor upsampling followed by convolutional operations, this representation is converted to a $256\times 256\times 3$ pixel image.
On the selector stream, it starts with a spatial projection of the embedding, which is then consecutively upsampled to be a series of feature maps which are concatenated with those feature maps in the generator, in order to offer guidance (from coarse to fine) at each scale.

Finally, the output of the generator is fed into a fixed pre-trained loss network to match the correlation statistics of the target texture using the visual features extracted at different layers of the loss network. We use the 19-layer VGG~\cite{VGG-2014} model as the loss network.

\subsection{Loss function}

We employ two loss functions, i.e., texture loss and diversity loss. The texture loss is computed between the synthesized result and the given texture to ensure that these two images share similar statistics and are perceptually similar.
The diversity loss is computed between outputs of the same texture (i.e., same input at selection unit) driven by different input noise vectors. The goal is to prevent the generator network from being trapped to a single degraded solution and to encourage the model to generate diversified results with large variations.

\begin{figure}[t]
\centering
\small
{
\begin{tabular}{c@{\hspace{0.01\linewidth}}c@{\hspace{0.01\linewidth}}c@{\hspace{0.01\linewidth}}c@{\hspace{0.01\linewidth}}c@{\hspace{0.01\linewidth}}c@{\hspace{0.01\linewidth}}c@{\hspace{0.01\linewidth}}c@{\hspace{0.01\linewidth}}c@{\hspace{0.01\linewidth}}c@{\hspace{0.01\linewidth}}c}

\includegraphics[width = .15\linewidth]{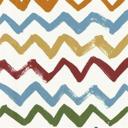} &
\includegraphics[width = .15\linewidth]{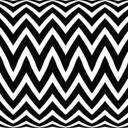} &
\includegraphics[width = .15\linewidth]{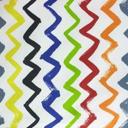} &

\hspace{1pt}\vrule\hspace{1pt}

\includegraphics[width = .15\linewidth]{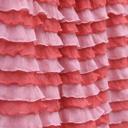} &
\includegraphics[width = .15\linewidth]{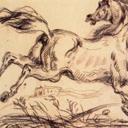} &
\includegraphics[width = .15\linewidth]{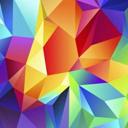} & \\

\includegraphics[width = .15\linewidth]{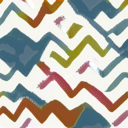} &
\includegraphics[width = .15\linewidth]{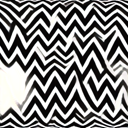} &
\includegraphics[width = .15\linewidth]{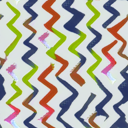} &

\hspace{1pt}\vrule\hspace{1pt}

\includegraphics[width = .15\linewidth]{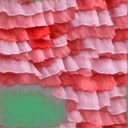} &
\includegraphics[width = .15\linewidth]{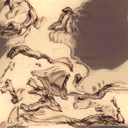} &
\includegraphics[width = .15\linewidth]{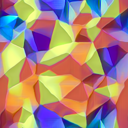} & \\

\includegraphics[width = .15\linewidth]{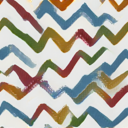} &
\includegraphics[width = .15\linewidth]{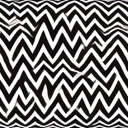} &
\includegraphics[width = .15\linewidth]{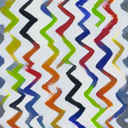} &

\hspace{1pt}\vrule\hspace{1pt}

\includegraphics[width = .15\linewidth]{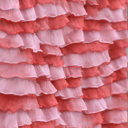} &
\includegraphics[width = .15\linewidth]{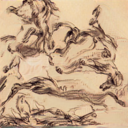} &
\includegraphics[width = .15\linewidth]{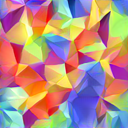} & \\

%{(a)}& {(b)}& {(c)}\\
\end{tabular}
}
\caption{Comparisons between using $G$ and mean subtracted $\overline{G}$. We show results of two 3-texture networks (left and right). Top: original textures, Middle: synthesized results using $G$ based texture loss, Bottom: synthesized results using $\overline{G}$ based texture loss.}
\label{fig:Gram_mean_subtraction}
\end{figure}

\paragraph{Texture loss.}
Similar to existing methods~\cite{GatysTexture-NIPS2015,Perceptual-ECCV2016,Texturenet-ICML2016},
the texture loss is based on the Gram matrix ($G$) difference of the feature maps in different layers of the loss network as
\begin{equation}\label{formula2_1}
L_{texture} = \|G_{gt} - G_{output}\|_{1}~,~~~ \\
G_{ij} = \sum \limits_{k} F_{ik}F_{jk}~,
\end{equation}
where each entry $G_{ij}$ in the Gram matrix is defined as the inner product of $F_{ik}$ and $F_{jk}$, which are vectorized activations of the $i$th (and $j$th) filter at position $k$ in the current layer of the loss network.
We use the activations at the $conv1\_ 1$, $conv2\_ 1$, $conv3\_ 1$, $conv4\_ 1$ and $conv5\_ 1$ layer of the VGG model.

The Gram matrix based texture loss has been shown demonstrated to effective for single texture synthesis.
However, for the purpose of multiple textures synthesis, we empirically
find that the original texture loss (defined as Eq.~\ref{formula2_1}) poses difficulty for the network to distinguish between textures and thus fails to synthesize them well.
In the middle row of Figure~\ref{fig:Gram_mean_subtraction}, we show a few examples of textures generated using the original texture loss in two experiments of synthesizing 3 textures with one network.
Note the obvious artifacts and color mixing problems in the synthesized results.
We attribute this problem to the large discrepancy in scale of the Gram matrices of different textures.

Motivated by this observation, we modify the original Gram matrix computation by subtracting the mean before calculating the inner product between two activations:
\begin{equation}\label{formula2}
\overline{G}_{ij} = \sum \limits_{k} (F_{ik}-\overline{F})(F_{jk}-\overline{F})~,
\end{equation}
where $\overline{F}$ is defined as the mean of all activations in the current layer of the loss network and the rest of terms remain the same with those in the definition of the Gram matrix~\eqref{formula2_1}.
Without re-centering the activations, we notice that during training the values of losses and gradients from different textures vary drastically, which suggests that the network is biased to learn the scale of Gram matrix, i.e., $\overline{F}$, instead of discriminating between them.
In the bottom row of Figure~\ref{fig:Gram_mean_subtraction}, we provide the same textures synthesized with the re-centered Gram matrix, which clearly shows improvements compared to the middle row.

\begin{figure}[t]
\centering
\small
{
\begin{tabular}{c@{\hspace{0.01\linewidth}}c@{\hspace{0.01\linewidth}}c@{\hspace{0.01\linewidth}}c@{\hspace{0.01\linewidth}}c@{\hspace{0.01\linewidth}}c@{\hspace{0.01\linewidth}}c@{\hspace{0.01\linewidth}}c@{\hspace{0.01\linewidth}}c@{\hspace{0.01\linewidth}}c@{\hspace{0.01\linewidth}}c}

\includegraphics[width = .125\linewidth]{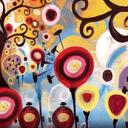} &

\hspace{1pt}\vrule\hspace{1pt}

\includegraphics[width = .125\linewidth]{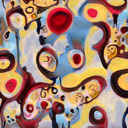} &
\includegraphics[width = .125\linewidth]{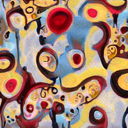} &
\includegraphics[width = .125\linewidth]{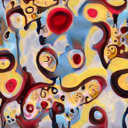} &

\hspace{1pt}\vrule\hspace{1pt}

\includegraphics[width = .125\linewidth]{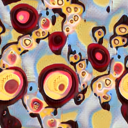} &
\includegraphics[width = .125\linewidth]{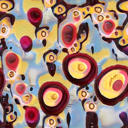} &
\includegraphics[width = .125\linewidth]{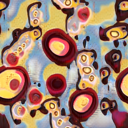} & \\

\includegraphics[width = .125\linewidth]{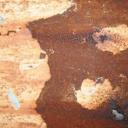} &

\hspace{1pt}\vrule\hspace{1pt}

\includegraphics[width = .125\linewidth]{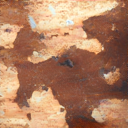} &
\includegraphics[width = .125\linewidth]{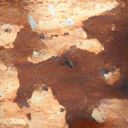} &
\includegraphics[width = .125\linewidth]{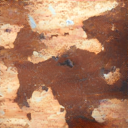} &

\hspace{1pt}\vrule\hspace{1pt}

%\hspace{1pt}\vrule\hspace{1pt}

%\includegraphics[width = .125\linewidth]{figs/Diversity/blank.jpg} &
\includegraphics[width = .125\linewidth]{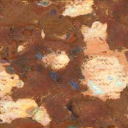} &
\includegraphics[width = .125\linewidth]{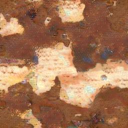} &
\includegraphics[width = .125\linewidth]{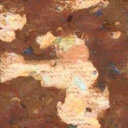} & \\

%{(a)}& {(b)}& {(c)}\\
\end{tabular}
}
\caption{Comparisons between without and with the diversity loss. Left: original textures, Middle: outputs (w/o diversity) under there different noise inputs, Right: outputs (w/ diversity) under the same set of different noise inputs.}
\label{fig:Diversity}
\end{figure}

\paragraph{Diversity loss.}

As mentioned above, one of the issues with existing feed-forward methods is being easily trapped to a degraded solution where it always outputs that are visually identical (sometimes with less satisfying repetitive patterns)~\cite{Diversity-NIPS2016}.
%
%The same goes to our model when trained with the texture loss only.
When trained only with the texture loss, the proposed network has the same issue.
We show several examples in the middle panel of Figure~\ref{fig:Diversity}.
The results under different noise input are nearly identical with subtle and unnoticeable difference in pixel values.
This is expected because the texture loss is designed to ensure all synthesized results to have the similar style with the given texture,
but does not enforce diversity among outputs.
In other words, each synthesized result is not correlated with the input noise.

In order to correlate the output with input noise, we design a diversity loss which explicitly measures the variation in visual appearance between the generated results under the same texture but different input noise.
Assume that there are $N$ input samples in a batch at each feed-forward pass, the generator will then emit $N$ outputs $\{P_{1},P_{2},...,P_{N}\}$.
Our diversity loss measures the visual difference between any pair of outputs $P_{i}$ and $P_{j}$ using visual features.
Let $\{Q_{1},Q_{2},...,Q_{N}\}$ be a random reordering of $\{P_{1},P_{2},...,P_{N}\}$, satisfying that $P_{i}\neq Q_{i}$.
In order to encourage the diversity in a higher level rather than lower level such as pixel shift, the diversity loss is computed between feature maps at the $conv4\_ 2$ layer of the loss network $\Phi$ as follows:
\begin{equation}\label{formula3}
L_{diversity} = \frac{1}{N} \sum_{i=1}^{N} \|\Phi (P_{i}) - \Phi (Q_{i})\|_{1},
\end{equation}
The results generated by our method with this diversity loss are shown in the right panel of Figure~\ref{fig:Diversity}.
While being perceptually similar, the results from our method contain rich variations.
Similar observations are found in~\cite{Diversity-NIPS2016} which also encourages diversity in generative model training by enlarging the distance among all samples within a batch on intermediate layers features, while our method achieves this with the diversity loss.

\begin{figure}[t]
\begin{center}
\includegraphics[width=9cm]{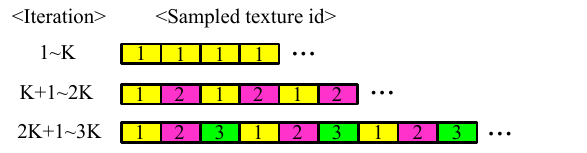}
\end{center}
\caption{Incremental training strategy. Each block represents an iteration and the number in it is the sampled texture id for this iteration (also the bit we set as $1$ in the selection unit). We use $K=1000$ in the experiments. }
\label{fig:Incremental60_strategy}
\end{figure}

\begin{figure}[t]
\begin{center}
\includegraphics[width=8cm]{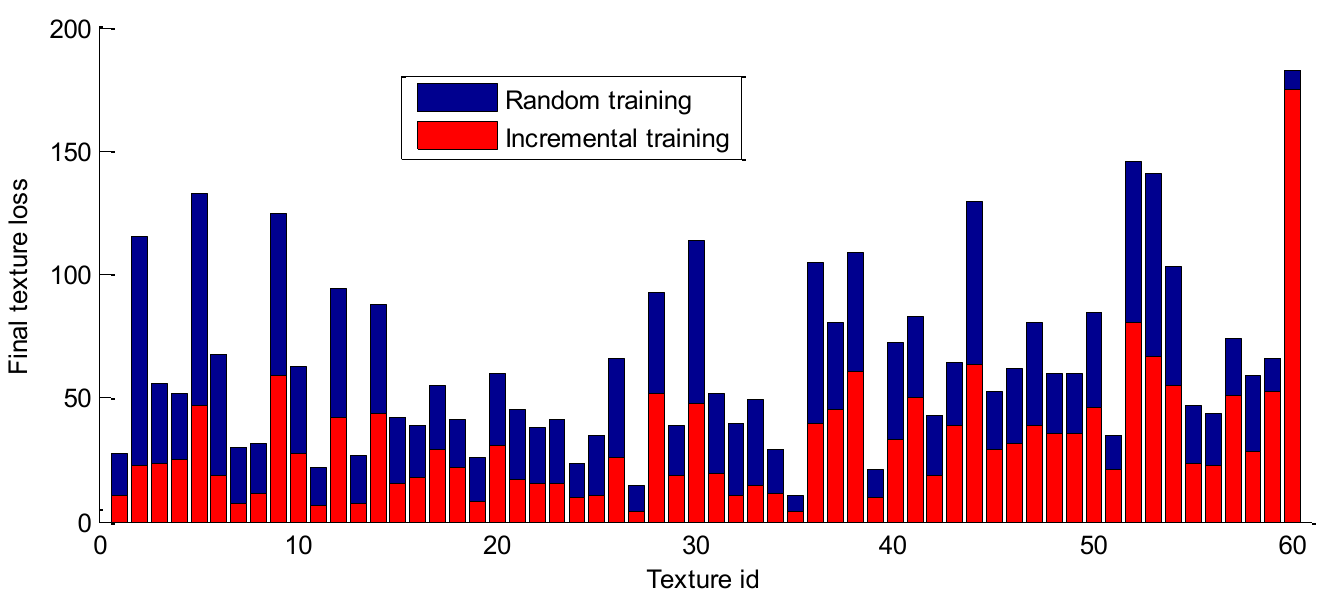}
\end{center}
\caption{Comparisons of the final texture loss when converged between the random and incremental training on 60-texture synthesis.}
\label{fig:Incremental60_1}
\end{figure}

The final loss function of our model is a combination of the texture loss and the diversity loss as shown in~\eqref{formula4}. As the goal is to minimize the texture loss and maximize the diversity loss, we use the coefficients $\alpha=1$, $\beta=-1$ in our experiments.
\begin{equation}\label{formula4}
L = \alpha L_{texture} + \beta L_{diversity},
\end{equation}

\subsection{Incremental training}

%We discuss our findings in training strategy in this section.
We discuss the training process for the proposed network with focus on how to sample a target texture among a set of predefined texture set.
%
%Our discussion mainly concerns about how to sample a target texture among a set of predefined texture set during the training.
%
%More specifically, our question is that should random sampling or some order of selecting target textures be adopted.
More specifically, we address the issue whether samples should randomly selected or in certain order in order to generate diversified textures.
Once a target texture is selected, we set the corresponding bit in the selection unit as $1$  and the corresponding texture is used to compute the texture loss.

\begin{figure*}[t]
\centering
\small
{
\begin{tabular}{c@{\hspace{0.01\linewidth}}c@{\hspace{0.01\linewidth}}c@{\hspace{0.01\linewidth}}c@{\hspace{0.01\linewidth}}c@{\hspace{0.01\linewidth}}c@{\hspace{0.01\linewidth}}c@{\hspace{0.01\linewidth}}c@{\hspace{0.01\linewidth}}c@{\hspace{0.01\linewidth}}c@{\hspace{0.01\linewidth}}c}

\includegraphics[width = .088\linewidth]{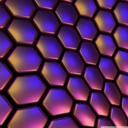} &
\includegraphics[width = .088\linewidth]{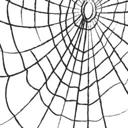} &
\includegraphics[width = .088\linewidth]{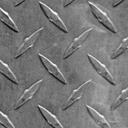} &
\includegraphics[width = .088\linewidth]{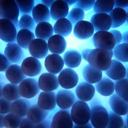} &
\includegraphics[width = .088\linewidth]{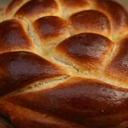} &
\includegraphics[width = .088\linewidth]{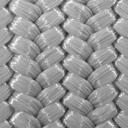} &
\includegraphics[width = .088\linewidth]{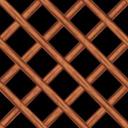} &
\includegraphics[width = .088\linewidth]{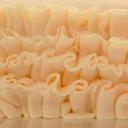} &
\includegraphics[width = .088\linewidth]{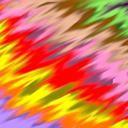} &
\includegraphics[width = .088\linewidth]{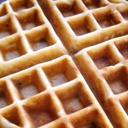} & \\

\includegraphics[width = .088\linewidth]{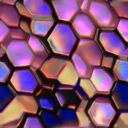} &
\includegraphics[width = .088\linewidth]{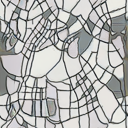} &
\includegraphics[width = .088\linewidth]{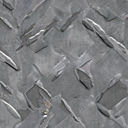} &
\includegraphics[width = .088\linewidth]{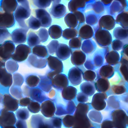} &
\includegraphics[width = .088\linewidth]{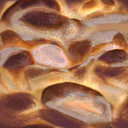} &
\includegraphics[width = .088\linewidth]{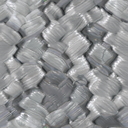} &
\includegraphics[width = .088\linewidth]{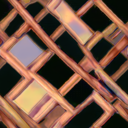} &
\includegraphics[width = .088\linewidth]{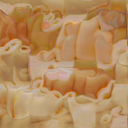} &
\includegraphics[width = .088\linewidth]{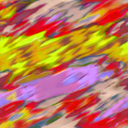} &
\includegraphics[width = .088\linewidth]{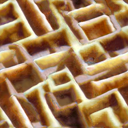} & \\

\includegraphics[width = .088\linewidth]{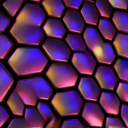} &
\includegraphics[width = .088\linewidth]{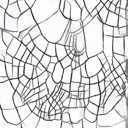} &
\includegraphics[width = .088\linewidth]{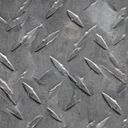} &
\includegraphics[width = .088\linewidth]{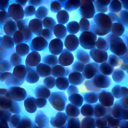} &
\includegraphics[width = .088\linewidth]{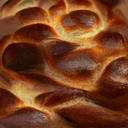} &
\includegraphics[width = .088\linewidth]{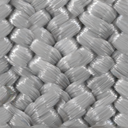} &
\includegraphics[width = .088\linewidth]{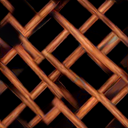} &
\includegraphics[width = .088\linewidth]{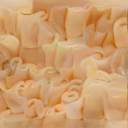} &
\includegraphics[width = .088\linewidth]{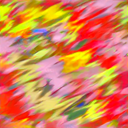} &
\includegraphics[width = .088\linewidth]{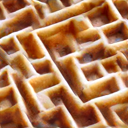} & \\

%{(a)}& {(b)}& {(c)}\\
\end{tabular}
}
\caption{Comparisons between random training and incremental training. Top: original textures, Middle: synthesis with random training, Bottom: synthesis with incremental training. The model is handling 60 textures in Figure~\ref{fig:Incremental60_1} and we show the synthesized results of 10 textures here.}
\label{fig:Incremental60_2}
\end{figure*}

%To our observation, we notice that the random sampling strategy typically yields inferior results and it becomes difficult to further push down texture losses for all texture images after certain number of iterations.
Empirically we find that the random sampling strategy typically yields inferior results and it becomes difficult to further push down texture losses
for all texture images after certain number of iterations.
We train a 60-texture network as an example and show the converged results (10 out of 60) with random sampling in the middle row of Figure~\ref{fig:Incremental60_2}.
The artifacts are clearly visible. Major patterns of each texture are captured, however the geometry is not well preserved (e.g., hexagons in the first texture), and colors are not well matched (i.e., mixing with colors from other textures).

We attribute this issue to the constant drastic change in the learning objective caused by random sampling within a diverse set of target textures.
In other words, although the network gains improvement toward a sampled texture at every iteration, the improvement is likely to be overwhelmed in the following iterations, where different textures are optimized.
As a consequence, the learning becomes less effective and eventually gets stuck to a bad local optimum.

Therefore we propose an incremental training strategy to help the learning to be more effective. Overall our incremental training strategy can be seen as a form of
curriculum learning.
There are two aspects in training the proposed network incrementally.
First, we do not teach the network to learn new tasks \emph{before} existing the network learns existing ones well.
That is, we start from learning one texture and gradually introduce new textures when the network can synthesize previous textures well.
Second, we ensure that the network does not \emph{forget} what is already learned.
Namely, we make sure that all the target textures fed to the network so far will still be sampled in future iterations, so that the network ``remembers''
how to synthesize them.

Specifically, in the first $K$ iterations, we keep setting the $1$st bit of the selection unit as $1$ to let the network fully focus on synthesizing Texture 1.
In the next $K$ iterations, Texture 2 is involved and we sample the bit from $1$ to $2$ in turn.
%
%The same goes to the rest of textures.  We use $K=1000$ in our experiments.
%
We repeat the same process to the other textures. We illustrate this procedure in Figure~\ref{fig:Incremental60_strategy}.
After all the textures are introduced to the network, we switch to the random sampling strategy until the training process converges.
In Figure~\ref{fig:Incremental60_1} and~\ref{fig:Incremental60_2}, we show the comparison
of both the final texture loss and synthesized visual results between the random and incremental training strategies
in the 60-texture network experiment.
Clearly the incremental training scheme leads to better convergence quantitatively and qualitatively.

\begin{figure}[t]
\centering
\small
\begin{tabular}{c@{\hspace{0.01\linewidth}}c@{\hspace{0.01\linewidth}}c@{\hspace{0.01\linewidth}}c@{\hspace{0.01\linewidth}}c@{\hspace{0.01\linewidth}}c@{\hspace{0.01\linewidth}}c}

\includegraphics[width=4cm]{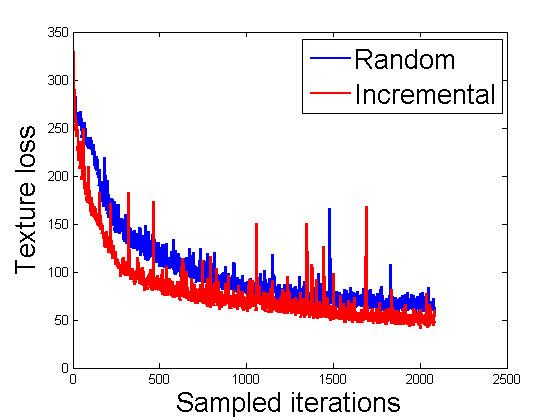} &
\includegraphics[width=4cm]{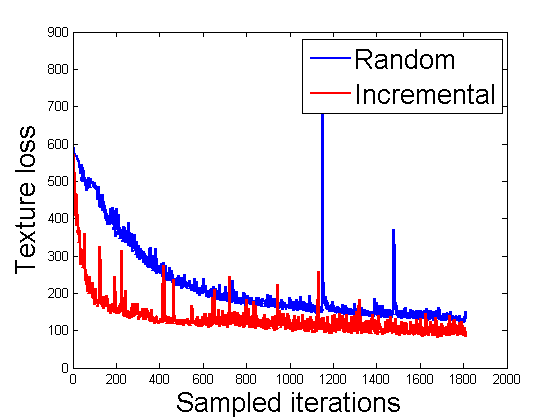}\\

{ (a) Texture 20} & { (b) Texture 30}\\

\includegraphics[width=4cm]{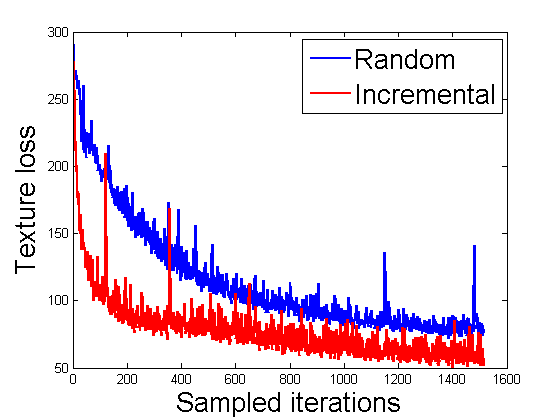} &
\includegraphics[width=4cm]{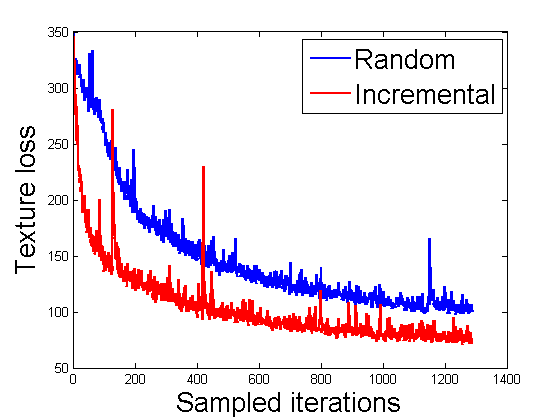}\\

 { (c) Texture 40}& { (d) Texture 50}\\
\end{tabular}
\caption{Comparisons between the random and incremental training on a single texture during a 60-texture network training. Note that some sudden drastic changes on the loss curve appear when a new texture is firstly involved which causes a short-term oscillation in the network.}
\label{fig:Incremental60_single}
\end{figure}

Interestingly, we observe that the network learns new textures faster as it sees more textures in later training stages.
To demonstrate that, we record the texture losses for each texture when it is sampled and show two examples in Figure~\ref{fig:Incremental60_single} when training the 60-texture network.
Take Texture 20 (Figure~\ref{fig:Incremental60_single}(a)) as an example, the network learned with incremental training quickly reaches lower losses compare to the one with random sampling strategy.
We hypothesize that the network benefits from the shared knowledge learned from Texture 1-19.
This conjecture is supported by later introduced textures (Figure~\ref{fig:Incremental60_single}(b-d)) where incremental training gets relatively faster at convergence as it learns more textures.

\begin{figure*}[t]
\centering
\small
{
\begin{tabular}{c@{\hspace{0.01\linewidth}}c@{\hspace{0.01\linewidth}}c@{\hspace{0.01\linewidth}}c@{\hspace{0.01\linewidth}}c@{\hspace{0.01\linewidth}}c@{\hspace{0.01\linewidth}}c@{\hspace{0.01\linewidth}}c@{\hspace{0.01\linewidth}}c@{\hspace{0.01\linewidth}}c@{\hspace{0.01\linewidth}}c}

\includegraphics[width = .085\linewidth]{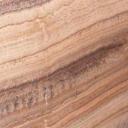} &
\includegraphics[width = .085\linewidth]{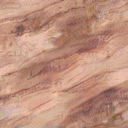} &

\hspace{1pt}\vrule\hspace{1pt}

\includegraphics[width = .085\linewidth]{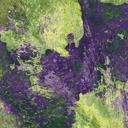} &
\includegraphics[width = .085\linewidth]{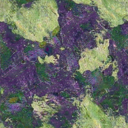} &

\hspace{1pt}\vrule\hspace{1pt}

\includegraphics[width = .085\linewidth]{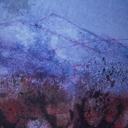} &
\includegraphics[width = .085\linewidth]{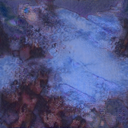} &

\hspace{1pt}\vrule\hspace{1pt}

\includegraphics[width = .085\linewidth]{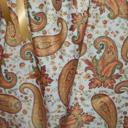} &
\includegraphics[width = .085\linewidth]{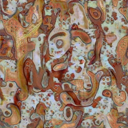} &

\hspace{1pt}\vrule\hspace{1pt}

\includegraphics[width = .085\linewidth]{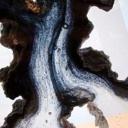} &
\includegraphics[width = .085\linewidth]{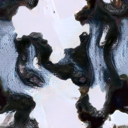} & \\

\includegraphics[width = .085\linewidth]{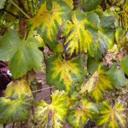} &
\includegraphics[width = .085\linewidth]{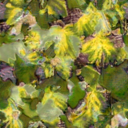} &

\hspace{1pt}\vrule\hspace{1pt}

\includegraphics[width = .085\linewidth]{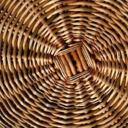} &
\includegraphics[width = .085\linewidth]{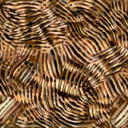} &

\hspace{1pt}\vrule\hspace{1pt}

\includegraphics[width = .085\linewidth]{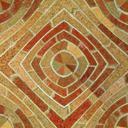} &
\includegraphics[width = .085\linewidth]{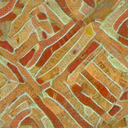} &

\hspace{1pt}\vrule\hspace{1pt}

\includegraphics[width = .085\linewidth]{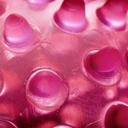} &
\includegraphics[width = .085\linewidth]{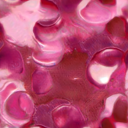} &

\hspace{1pt}\vrule\hspace{1pt}

\includegraphics[width = .085\linewidth]{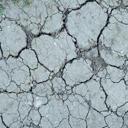} &
\includegraphics[width = .085\linewidth]{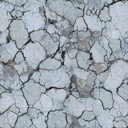} & \\

\includegraphics[width = .085\linewidth]{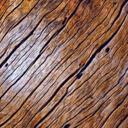} &
\includegraphics[width = .085\linewidth]{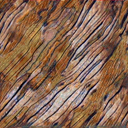} &

\hspace{1pt}\vrule\hspace{1pt}

\includegraphics[width = .085\linewidth]{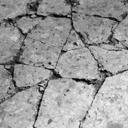} &
\includegraphics[width = .085\linewidth]{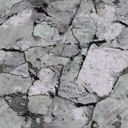} &

\hspace{1pt}\vrule\hspace{1pt}

\includegraphics[width = .085\linewidth]{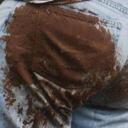} &
\includegraphics[width = .085\linewidth]{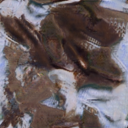} &

\hspace{1pt}\vrule\hspace{1pt}

\includegraphics[width = .085\linewidth]{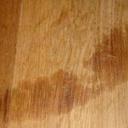} &
\includegraphics[width = .085\linewidth]{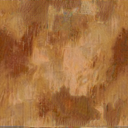} &

\hspace{1pt}\vrule\hspace{1pt}

\includegraphics[width = .085\linewidth]{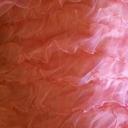} &
\includegraphics[width = .085\linewidth]{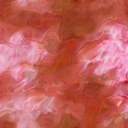} & \\

\includegraphics[width = .085\linewidth]{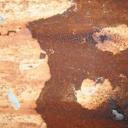} &
\includegraphics[width = .085\linewidth]{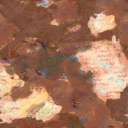} &

\hspace{1pt}\vrule\hspace{1pt}

\includegraphics[width = .085\linewidth]{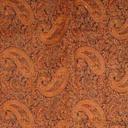} &
\includegraphics[width = .085\linewidth]{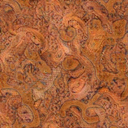} &

\hspace{1pt}\vrule\hspace{1pt}

\includegraphics[width = .085\linewidth]{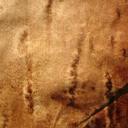} &
\includegraphics[width = .085\linewidth]{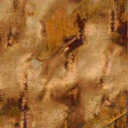} &

\hspace{1pt}\vrule\hspace{1pt}

\includegraphics[width = .085\linewidth]{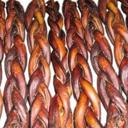} &
\includegraphics[width = .085\linewidth]{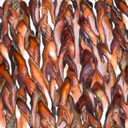} &

\hspace{1pt}\vrule\hspace{1pt}

\includegraphics[width = .085\linewidth]{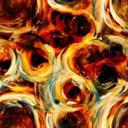} &
\includegraphics[width = .085\linewidth]{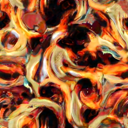} & \\

%{(a)}& {(b)}& {(c)}\\
\end{tabular}
}
\caption{Synthesized results of a 300-texture network. In each panel, Left: original texture, Right: synthesized result. We show results of 20 (out of 300) textures as examples here. For each texture, we only show one synthesized result. % and more diverse results are included in the supplementary material.
}
\label{fig:Texture300}
\end{figure*}

\section{Experimental Results}

In this section, we present extensive experimental results to demonstrate the effectiveness of our algorithm.
We experiment with synthesizing a large number of textures using a single network and then
show that our model is able to generate diverse outputs and create new textures by linear interpolation.
%
%Furthermore, we extend our multi-texture synthesis framework to multi-style transfer, and study how incremental training improves style transfer.
%
%More results can be found in the appendix.

\subsection{Multi-texture synthesis}

In addition to the 60-texture network trained for illustration purpose in Section 3,
we experiment with a larger 300-texture network to further validate the robustness and scalability of our model.
We map the 300-dimensional selection unit to a 128-dimensional embedding and use a 5-dimensional noise vector.
The network is trained with both texture and diversity loss under the incremental training strategy. Texture images used in our experiments are from the Describable Textures Dataset (DTD)~\cite{DTDtexture-CVPR2016}.
Figure~\ref{fig:Texture300} shows the synthesized results of 20 textures.
%
%More results are presented in the supplementary material.
%We only present one result for each texture but the network is able to generate diverse outputs conditioned on different noise inputs, as shown in Figure~\ref{fig:Diversity}.
%
%More results are presented in the supplementary material.

%Additionally, our model is more efficient at memory overhead, which could be an important factor for practical usage.
%
%The size of our 300-texture network is around 106MB while the total size of 300 separate networks based on TextureNet~\cite{Texturenet-ICML2016} is around 330MB.

\begin{figure}[t]
\centering
\small
{
\begin{tabular}{c@{\hspace{0.01\linewidth}}c@{\hspace{0.01\linewidth}}c@{\hspace{0.01\linewidth}}c@{\hspace{0.01\linewidth}}c@{\hspace{0.01\linewidth}}c@{\hspace{0.01\linewidth}}c@{\hspace{0.01\linewidth}}c@{\hspace{0.01\linewidth}}c@{\hspace{0.01\linewidth}}c@{\hspace{0.01\linewidth}}c}

\includegraphics[width = .125\linewidth]{figs/Diversity/1.jpg} &

\hspace{1pt}\vrule\hspace{1pt}

\includegraphics[width = .125\linewidth]{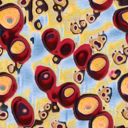} &
\includegraphics[width = .125\linewidth]{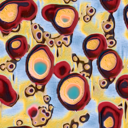} &
\includegraphics[width = .125\linewidth]{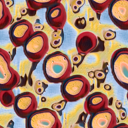} &

\hspace{1pt}\vrule\hspace{1pt}

\includegraphics[width = .125\linewidth]{figs/Diversity/1_1.png} &
\includegraphics[width = .125\linewidth]{figs/Diversity/1_2.png} &
\includegraphics[width = .125\linewidth]{figs/Diversity/1_3.png} & \\

\includegraphics[width = .125\linewidth]{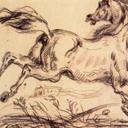} &

\hspace{1pt}\vrule\hspace{1pt}

\includegraphics[width = .125\linewidth]{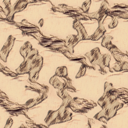} &
\includegraphics[width = .125\linewidth]{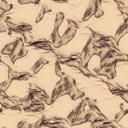} &
\includegraphics[width = .125\linewidth]{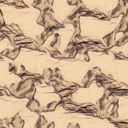} &

\hspace{1pt}\vrule\hspace{1pt}

\includegraphics[width = .125\linewidth]{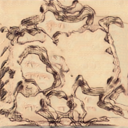} &
\includegraphics[width = .125\linewidth]{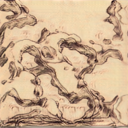} &
\includegraphics[width = .125\linewidth]{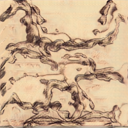} & \\

\includegraphics[width = .125\linewidth]{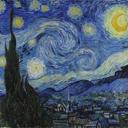} &

\hspace{1pt}\vrule\hspace{1pt}

\includegraphics[width = .125\linewidth]{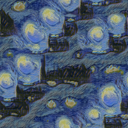} &
\includegraphics[width = .125\linewidth]{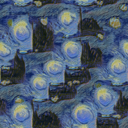} &
\includegraphics[width = .125\linewidth]{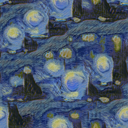} &

\hspace{1pt}\vrule\hspace{1pt}

\includegraphics[width = .125\linewidth]{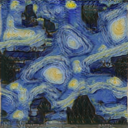} &
\includegraphics[width = .125\linewidth]{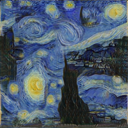} &
\includegraphics[width = .125\linewidth]{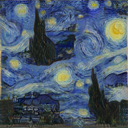} & \\

%{(a)}& {(b)}& {(c)}\\
\end{tabular}
}
\caption{Comparisons of diverse synthesized results between the TextureNet~\cite{Texturenet-ICML2016} (middle) and our model (right).}
\label{fig:Diversity_vs_TextureNet}
\end{figure}

\begin{figure*}[t]
\centering
\small
{
\begin{tabular}{c@{\hspace{0.01\linewidth}}c@{\hspace{0.01\linewidth}}c@{\hspace{0.01\linewidth}}c@{\hspace{0.01\linewidth}}c@{\hspace{0.01\linewidth}}c@{\hspace{0.01\linewidth}}c@{\hspace{0.01\linewidth}}c@{\hspace{0.01\linewidth}}c@{\hspace{0.01\linewidth}}c@{\hspace{0.01\linewidth}}c}

\includegraphics[width = .087\linewidth]{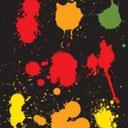} &

\hspace{1pt}\vrule\hspace{1pt}

\includegraphics[width = .087\linewidth]{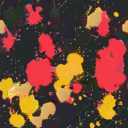} &
\includegraphics[width = .087\linewidth]{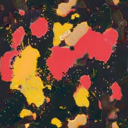} &
\includegraphics[width = .087\linewidth]{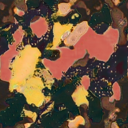} &
\includegraphics[width = .087\linewidth]{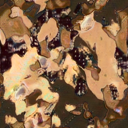} &
\includegraphics[width = .087\linewidth]{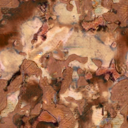} &
\includegraphics[width = .087\linewidth]{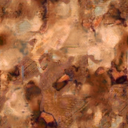} &
\includegraphics[width = .087\linewidth]{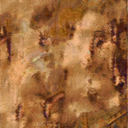} &
\includegraphics[width = .087\linewidth]{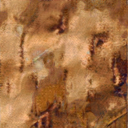} &

\hspace{1pt}\vrule\hspace{1pt}

\includegraphics[width = .087\linewidth]{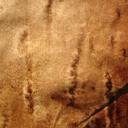} & \\

\includegraphics[width = .087\linewidth]{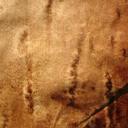} &

\hspace{1pt}\vrule\hspace{1pt}

\includegraphics[width = .087\linewidth]{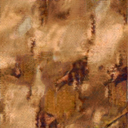} &
\includegraphics[width = .087\linewidth]{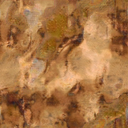} &
\includegraphics[width = .087\linewidth]{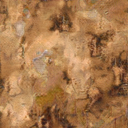} &
\includegraphics[width = .087\linewidth]{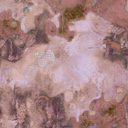} &
\includegraphics[width = .087\linewidth]{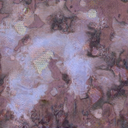} &
\includegraphics[width = .087\linewidth]{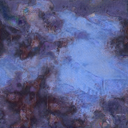} &
\includegraphics[width = .087\linewidth]{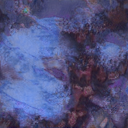} &
\includegraphics[width = .087\linewidth]{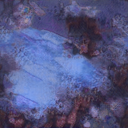} &

\hspace{1pt}\vrule\hspace{1pt}

\includegraphics[width = .087\linewidth]{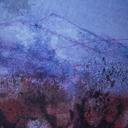} & \\

%{(a)}& {(b)}& {(c)}\\
\end{tabular}
}
\caption{Texture interpolation (or transition) with the 300-texture network. Top: Texture 20 to Texture 19, Bottom: Texture 19 to Texture 12. Images in the leftmost and rightmost are original textures.}
\label{fig:Interpolation}
\end{figure*}

\subsection{Diversity}

By sampling different noise in the noise vector, our network can generate diverse synthesized results for each texture. Existing single-texture networks~\cite{Texturenet-ICML2016} can also generate diversity to a certain extent.
However, the diversity is still limited because their network is trained with the texture loss only.
The diversity in~\cite{Texturenet-ICML2016} is mainly enforced by injecting multiple noise maps at different scales (from $8\times 8$ to $256\times 256$).
Without explicit constraints to push diversity, such a huge variation will be reduced or absorbed by the network, which still leads to limited diversity in outputs.
We compare the diverse outputs of our model and~\cite{Texturenet-ICML2016} in Figure~\ref{fig:Diversity_vs_TextureNet}.
Note that the common diagonal layout is shared across different results of~\cite{Texturenet-ICML2016}, which causes unsatisfying visual experience.
In contrast, our method achieves diversity in a more natural and flexible manner.
With the diversity loss, our model enables diverse outputs with low dimensional noise input, which gives us the ability to generate continuous transition between those outputs.
%
%We present a set of videos that show the transition of synthesized textures in the supplementary material.

\subsection{Interpolation}

Equipped with a selection unit and a learned $M$-texture network, we can interpolate between bits at test time to create \emph{new} textures or generate smooth transitions between textures.
We show two examples of interpolation with our previously trained 300-texture network in Figure~\ref{fig:Interpolation}.
For example in the top row of Figure~\ref{fig:Interpolation}, we start from Texture 20 and drive the synthesis towards Texture 19.
This is carried out by gradually decreasing the weight in the 20th bit and increasing the weight in the 19th bit with the rest bits all set as zero.
Such a smooth transition indicates that our generation can go along a continuous space.

\begin{figure}[t]
\centering
\small
{
\begin{tabular}{c@{\hspace{0.01\linewidth}}c@{\hspace{0.01\linewidth}}c@{\hspace{0.01\linewidth}}c@{\hspace{0.01\linewidth}}c@{\hspace{0.01\linewidth}}c@{\hspace{0.01\linewidth}}c@{\hspace{0.01\linewidth}}c@{\hspace{0.01\linewidth}}c@{\hspace{0.01\linewidth}}c@{\hspace{0.01\linewidth}}c}

\ignore{
\includegraphics[width = .23\linewidth]{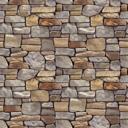} &
\includegraphics[width = .23\linewidth, height = .23\linewidth]{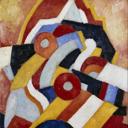} &
\includegraphics[width = .23\linewidth]{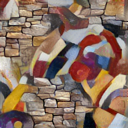} &
\includegraphics[width = .23\linewidth]{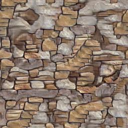} &\\
}

\includegraphics[width = .23\linewidth]{Rebuttal/Rebuttal_figs/Interpolation/t1.jpg} &
\includegraphics[width = .23\linewidth, height = .23\linewidth]{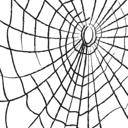} &
\includegraphics[width = .23\linewidth]{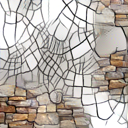} &
\includegraphics[width = .23\linewidth]{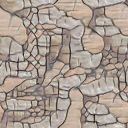} &\\

\includegraphics[width = .23\linewidth]{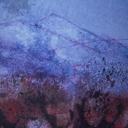} &
\includegraphics[width = .23\linewidth]{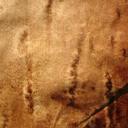} &
\includegraphics[width = .23\linewidth]{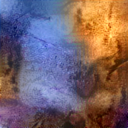} &
\includegraphics[width = .23\linewidth]{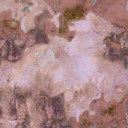} & \\

{Texture I}& {Texture II}& {\cite{GatysTransfer-CVPR2016}}& {Ours}\\

\end{tabular}
}
%\vspace{-2mm}
\caption{Interpolation comparison between~\cite{GatysTransfer-CVPR2016} and our method.}
\label{fig:Interpolation_vs_Deepart1}
%\vspace{-1.5em}
\end{figure}

The method in~\cite{GatysTransfer-CVPR2016} is also able to synthesize the interpolated result of two textures.
In~\cite{GatysTransfer-CVPR2016}, if we denote $G_1$ and $G_2$ as the Gram matrix of two textures, the interpolated texture is generated by matching some intermediate Gram matrix $a\times G_1 + (1-a)\times G_2$ through optimization (e.g., a=0.5).
We show the interpolation comparison between~\cite{GatysTransfer-CVPR2016} and our method in Figure~\ref{fig:Interpolation_vs_Deepart1}. It is observed that the results by~\cite{GatysTransfer-CVPR2016} are simply overlaid by two textures while our method generates
new textural effects.

\subsection{Extension to multi-style transfer}

We extend the idea of multi-texture synthesis to the multi-style transfer for image stylization.
Given a style image and a content image, the style transfer aims at synthesizing an image that preserves the global content while transferring the colors and local structures from the style image.
For presentation clarity, we will use the term \emph{style} instead of the \emph{texture}.

\begin{figure}[t]
\begin{center}
\includegraphics[width=8cm]{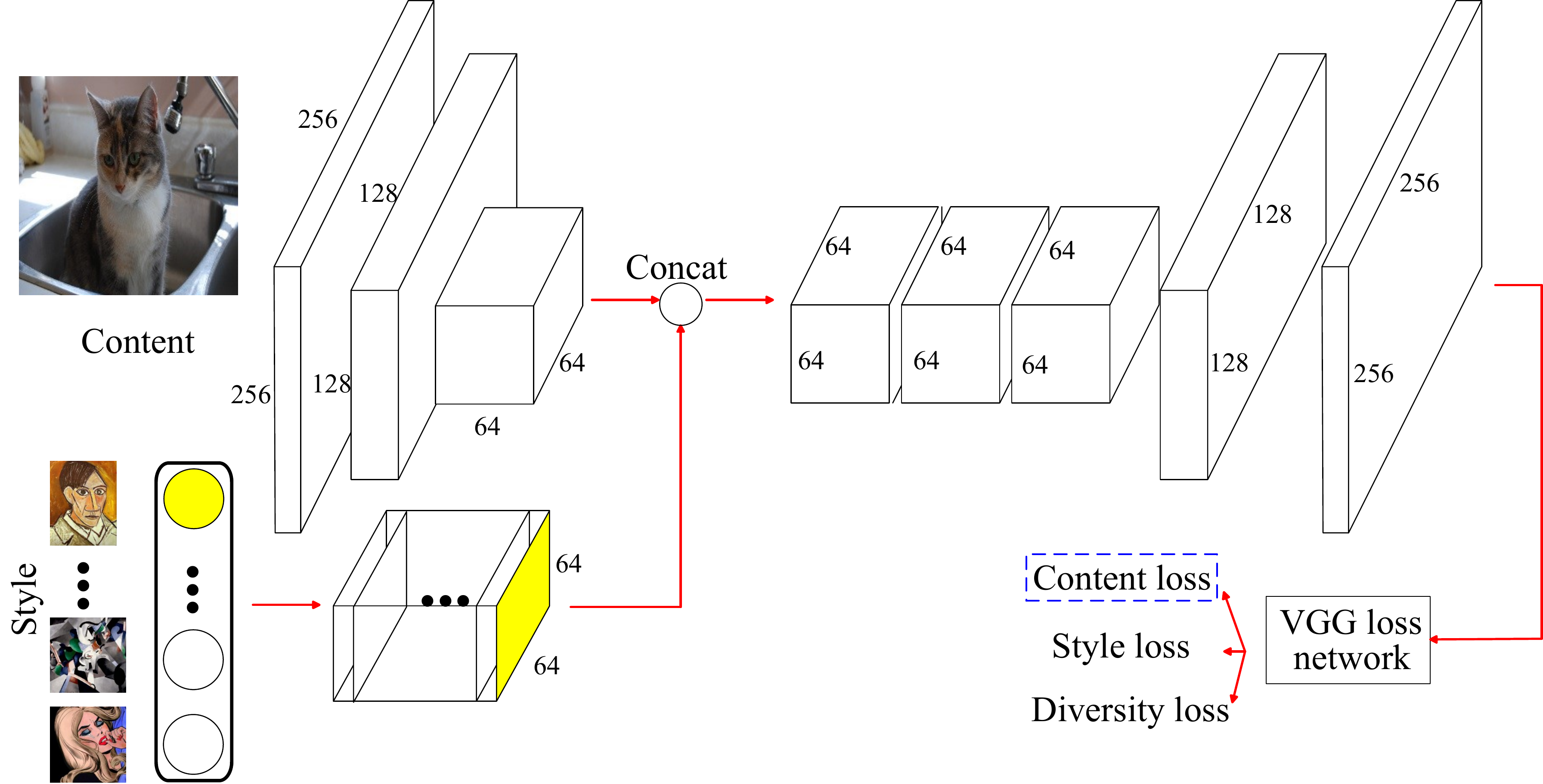}
\end{center}
\caption{Architecture of the multi-style transfer network.}
\label{fig:framework_transfer}
\end{figure}

\begin{figure*}[t]
\centering
\small
{
\begin{tabular}{c@{\hspace{0.01\linewidth}}c@{\hspace{0.01\linewidth}}c@{\hspace{0.01\linewidth}}c@{\hspace{0.01\linewidth}}c@{\hspace{0.01\linewidth}}c@{\hspace{0.01\linewidth}}c@{\hspace{0.01\linewidth}}c@{\hspace{0.01\linewidth}}c@{\hspace{0.01\linewidth}}c@{\hspace{0.01\linewidth}}c}

\includegraphics[width = .1\linewidth]{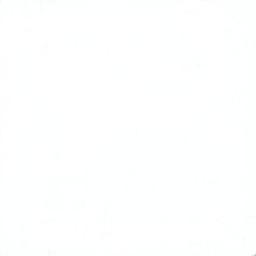} &
\includegraphics[width = .1\linewidth]{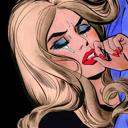} &
\includegraphics[width = .1\linewidth]{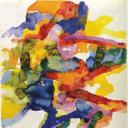} &
\includegraphics[width = .1\linewidth]{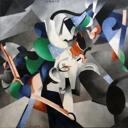} &
\includegraphics[width = .1\linewidth]{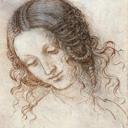} &
\includegraphics[width = .1\linewidth]{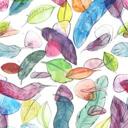} &
\includegraphics[width = .1\linewidth]{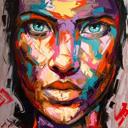} &
\includegraphics[width = .1\linewidth]{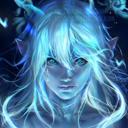} &
\includegraphics[width = .1\linewidth]{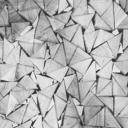} &\\

\includegraphics[width = .1\linewidth, height = .1\linewidth]{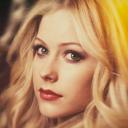} &
\includegraphics[width = .1\linewidth]{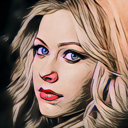} &
\includegraphics[width = .1\linewidth]{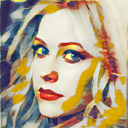} &
\includegraphics[width = .1\linewidth]{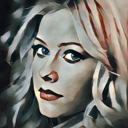} &
\includegraphics[width = .1\linewidth]{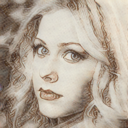} &
\includegraphics[width = .1\linewidth]{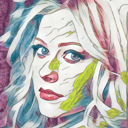} &
\includegraphics[width = .1\linewidth]{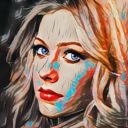} &
\includegraphics[width = .1\linewidth]{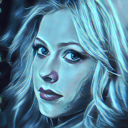} &
\includegraphics[width = .1\linewidth]{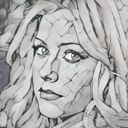} & \\

\includegraphics[width = .1\linewidth]{figs/Transfer/blank.jpg} &
\includegraphics[width = .1\linewidth]{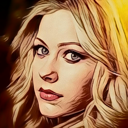} &
\includegraphics[width = .1\linewidth]{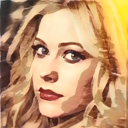} &
\includegraphics[width = .1\linewidth]{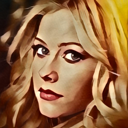} &
\includegraphics[width = .1\linewidth]{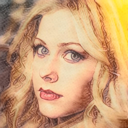} &
\includegraphics[width = .1\linewidth]{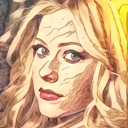} &
\includegraphics[width = .1\linewidth]{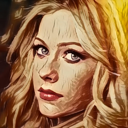} &
\includegraphics[width = .1\linewidth]{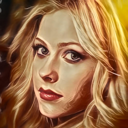} &
\includegraphics[width = .1\linewidth]{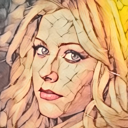} & \\

%{(a)}& {(b)}& {(c)}\\
\end{tabular}
}
\caption{Transferred results on \emph{test} content images of a 16-style network. We show results of 8 (out of 16) styles as examples. Top: style images, Leftmost: content image, Middle: transferred results. Bottom: color-independent transferred results.
}
\label{fig:Transfer16_results_selection_map1}
\end{figure*}

\begin{figure*}[h!]
\centering
\small
{
\begin{tabular}{c@{\hspace{0.01\linewidth}}c@{\hspace{0.01\linewidth}}c@{\hspace{0.01\linewidth}}c@{\hspace{0.01\linewidth}}c@{\hspace{0.01\linewidth}}c@{\hspace{0.01\linewidth}}c@{\hspace{0.01\linewidth}}c@{\hspace{0.01\linewidth}}c@{\hspace{0.01\linewidth}}c@{\hspace{0.01\linewidth}}c}

\includegraphics[height = .12\linewidth]{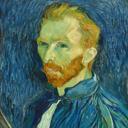} &
\includegraphics[height = .12\linewidth]{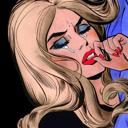} &
\includegraphics[height = .12\linewidth]{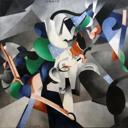} &
\includegraphics[height = .12\linewidth]{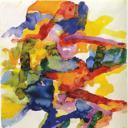} &
\includegraphics[height = .12\linewidth]{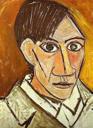} & \\

{Style 1}& {Style 2}& {Style 3}& {Style 4}& {Style 5}\\

\includegraphics[height = .12\linewidth]{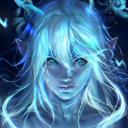} &
\includegraphics[height = .12\linewidth]{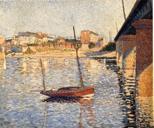} &
\includegraphics[height = .12\linewidth]{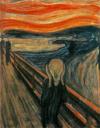} &
\includegraphics[height = .12\linewidth]{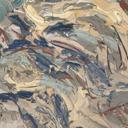} &
\includegraphics[height = .12\linewidth]{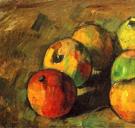} & \\

{Style 6}& {Style 7}& {Style 8}& {Style 9}& {Style 10}\\

\includegraphics[height = .12\linewidth]{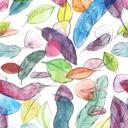} &
\includegraphics[height = .12\linewidth]{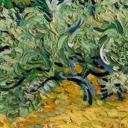} &
\includegraphics[height = .12\linewidth]{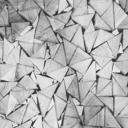} &
\includegraphics[height = .12\linewidth]{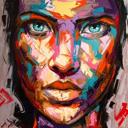} &
\includegraphics[height = .12\linewidth]{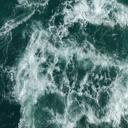} &
\includegraphics[height = .12\linewidth]{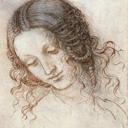} & \\

{Style 11}& {Style 12}& {Style 13}& {Style 14}& {Style 15}& {Style 16}\\

\includegraphics[height = .12\linewidth]{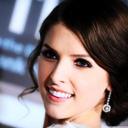} &
\includegraphics[height = .12\linewidth]{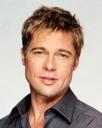} &
\includegraphics[height = .12\linewidth]{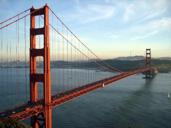} & \\

{Content 1}& {Content 2}& {Content 3}\\

\end{tabular}
}
\caption{16 styles used for our 16-style transfer model training and 3 test content images (bottom). All of them are resized to 512 but the aspect ratio is kept. Note that as our multi-transfer model is fully convolutional, it is able to handle the content image of arbitrary sizes.}
\label{fig:Transfer16_appendix_style}
\end{figure*}

\begin{figure*}[h!]
\centering
\small
{
\begin{tabular}{c@{\hspace{0.01\linewidth}}c@{\hspace{0.01\linewidth}}c@{\hspace{0.01\linewidth}}c@{\hspace{0.01\linewidth}}c@{\hspace{0.01\linewidth}}c@{\hspace{0.01\linewidth}}c@{\hspace{0.01\linewidth}}c@{\hspace{0.01\linewidth}}c@{\hspace{0.01\linewidth}}c@{\hspace{0.01\linewidth}}c}

\includegraphics[height = .15\linewidth]{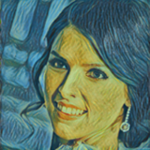} &
\includegraphics[height = .15\linewidth]{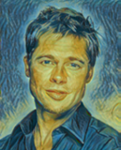} &
\includegraphics[height = .15\linewidth]{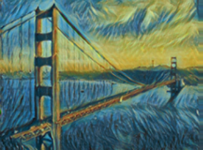} &

\hspace{1pt}\vrule\hspace{1pt}

\includegraphics[height = .15\linewidth]{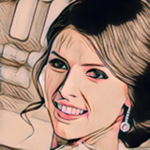} &
\includegraphics[height = .15\linewidth]{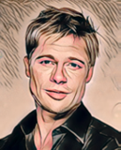} &
\includegraphics[height = .15\linewidth]{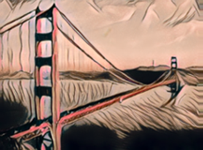} & \\

\includegraphics[height = .15\linewidth]{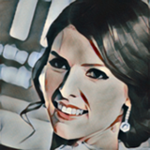} &
\includegraphics[height = .15\linewidth]{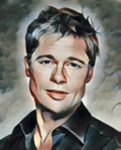} &
\includegraphics[height = .15\linewidth]{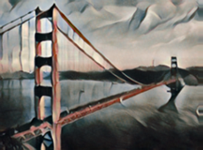} &

\hspace{1pt}\vrule\hspace{1pt}

\includegraphics[height = .15\linewidth]{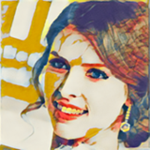} &
\includegraphics[height = .15\linewidth]{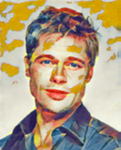} &
\includegraphics[height = .15\linewidth]{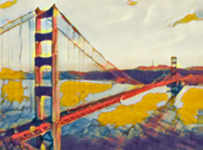} & \\

\includegraphics[height = .15\linewidth]{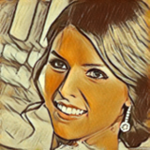} &
\includegraphics[height = .15\linewidth]{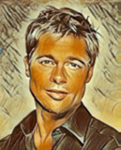} &
\includegraphics[height = .15\linewidth]{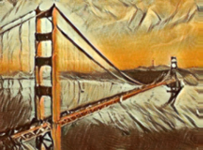} &

\hspace{1pt}\vrule\hspace{1pt}

\includegraphics[height = .15\linewidth]{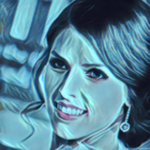} &
\includegraphics[height = .15\linewidth]{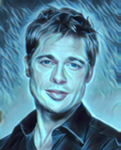} &
\includegraphics[height = .15\linewidth]{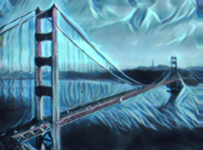} & \\

\includegraphics[height = .15\linewidth]{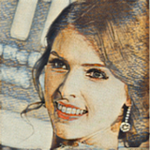} &
\includegraphics[height = .15\linewidth]{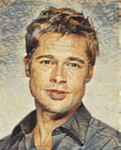} &
\includegraphics[height = .15\linewidth]{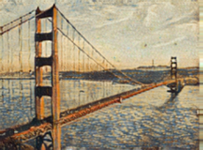} &

\hspace{1pt}\vrule\hspace{1pt}

\includegraphics[height = .15\linewidth]{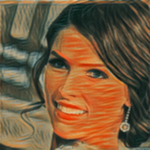} &
\includegraphics[height = .15\linewidth]{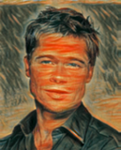} &
\includegraphics[height = .15\linewidth]{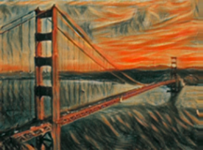} & \\

\includegraphics[height = .15\linewidth]{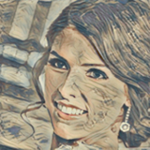} &
\includegraphics[height = .15\linewidth]{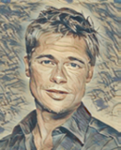} &
\includegraphics[height = .15\linewidth]{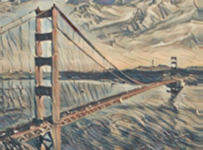} &

\hspace{1pt}\vrule\hspace{1pt}

\includegraphics[height = .15\linewidth]{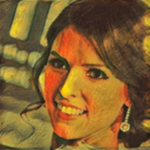} &
\includegraphics[height = .15\linewidth]{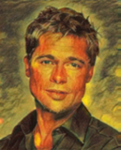} &
\includegraphics[height = .15\linewidth]{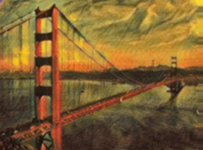} & \\

\includegraphics[height = .15\linewidth]{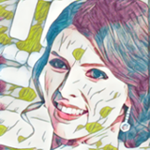} &
\includegraphics[height = .15\linewidth]{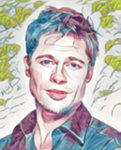} &
\includegraphics[height = .15\linewidth]{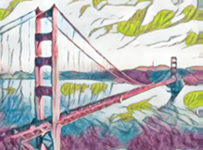} &

\hspace{1pt}\vrule\hspace{1pt}

\includegraphics[height = .15\linewidth]{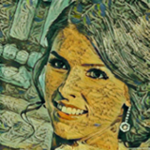} &
\includegraphics[height = .15\linewidth]{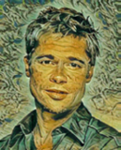} &
\includegraphics[height = .15\linewidth]{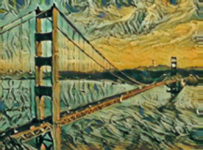} & \\

\includegraphics[height = .15\linewidth]{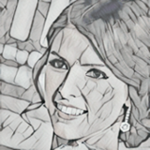} &
\includegraphics[height = .15\linewidth]{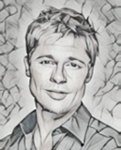} &
\includegraphics[height = .15\linewidth]{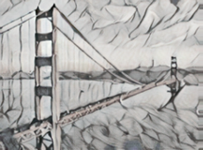} &

\hspace{1pt}\vrule\hspace{1pt}

\includegraphics[height = .15\linewidth]{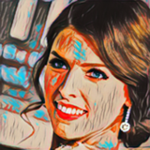} &
\includegraphics[height = .15\linewidth]{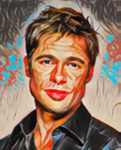} &
\includegraphics[height = .15\linewidth]{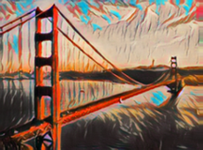} & \\

\includegraphics[height = .15\linewidth]{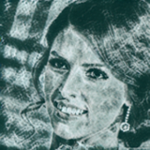} &
\includegraphics[height = .15\linewidth]{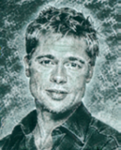} &
\includegraphics[height = .15\linewidth]{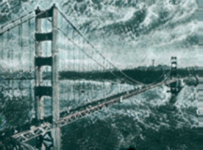} &

\hspace{1pt}\vrule\hspace{1pt}

\includegraphics[height = .15\linewidth]{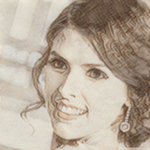} &
\includegraphics[height = .15\linewidth]{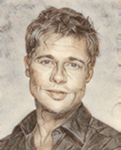} &
\includegraphics[height = .15\linewidth]{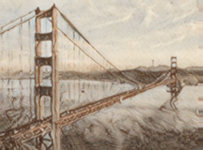} & \\

\end{tabular}
}
\caption{Transferred results of Style 1-16.}
\label{fig:Transfer16_appendix_result}
\end{figure*}

The network architecture is shown in Figure~\ref{fig:framework_transfer}.
We use an autoencoder network similar to~\cite{Perceptual-ECCV2016} and incorporate our idea of introducing a selection unit to handle the transferring of different styles.
%Therefore we mainly modify the synthesis model by introducing a stream of input from the content image and a content loss to help preserve the global arrangement.
%
More specifically, for each bit in the selection unit, we generate a corresponding noise map (e.g., from the uniform distribution) and concatenate these maps with the encoded features from the content, which are then decoded to the transferred result. When one style is selected, only the noise map that corresponds to it is randomly initialized while other noise maps are set to zero.
The content loss is computed as the feature differences between the transferred result and the content at the $conv4\_ 2$ layer of the VGG model
as in~\cite{GatysTransfer-CVPR2016}.
The style loss and diversity loss are defined in the same way as those in texture synthesis.
We train a 16-style transfer network and show the transferred results in Figure~\ref{fig:Transfer16_results_selection_map1}.
We also perform the luminance-only transfer as in~\cite{Gatys-PreservingColor-2016,Gatys2016-control} and show the color-independent transferred results in the bottom row of Figure~\ref{fig:Transfer16_results_selection_map1}.
%
%Note that as our multi-transfer model is fully convolutional, it is able to handle the content image of arbitrary sizes.
%
More results can be found in Figure~\ref{fig:Transfer16_appendix_style}-\ref{fig:Transfer16_appendix_result}.

\begin{figure}[t]
\centering
\small
{
\begin{tabular}{c@{\hspace{0.01\linewidth}}c@{\hspace{0.01\linewidth}}c@{\hspace{0.01\linewidth}}c@{\hspace{0.01\linewidth}}c@{\hspace{0.01\linewidth}}c@{\hspace{0.01\linewidth}}c@{\hspace{0.01\linewidth}}c@{\hspace{0.01\linewidth}}c@{\hspace{0.01\linewidth}}c@{\hspace{0.01\linewidth}}c}

\includegraphics[width = .148\linewidth]{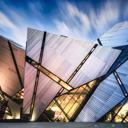} &

\hspace{1pt}\vrule\hspace{1pt}

\includegraphics[width = .148\linewidth]{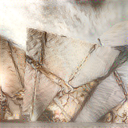} &
\includegraphics[width = .148\linewidth]{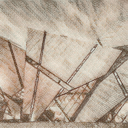} &
\includegraphics[width = .148\linewidth]{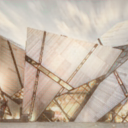} &
\includegraphics[width = .148\linewidth]{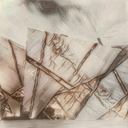} &

\hspace{1pt}\vrule\hspace{1pt}

\includegraphics[width = .148\linewidth]{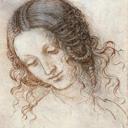} & \\

%{}& {2.21e5}& {2.46e5}& {2.39e5}& {2.40e5}& {}\\

\includegraphics[width = .148\linewidth]{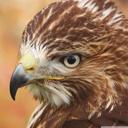} &

\hspace{1pt}\vrule\hspace{1pt}

\includegraphics[width = .148\linewidth]{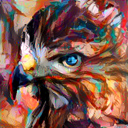} &
\includegraphics[width = .148\linewidth]{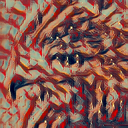} &
\includegraphics[width = .148\linewidth]{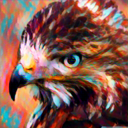} &
\includegraphics[width = .148\linewidth]{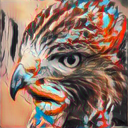} &

\hspace{1pt}\vrule\hspace{1pt}

\includegraphics[width = .148\linewidth]{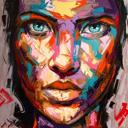} & \\

%{Content}& {2.24e6}& {2.23e6}& {2.27e6}& {2.25e6}& {Style}\\

{Content}& {\cite{GatysTransfer-CVPR2016}}& {\cite{Texturenet-ICML2016}}& {\cite{Perceptual-ECCV2016}}& {Ours}& {Style}\\

\end{tabular}
}

\caption{Comparison of style transfer results between existing methods and ours.}% Our results are obtained from our 16-style transfer network.}
\label{fig:Transfer_comparison_with_other_methods}
\end{figure}

In addition, we compare our multi-style transfer model with existing methods in Figure~\ref{fig:Transfer_comparison_with_other_methods}.
We adjust the style weight such that all methods have similar transferring effects. It clearly shows that our multi-style transfer model achieves improved or comparable results.

\begin{figure*}[t]
\centering
\small
{
\begin{tabular}{c@{\hspace{0.01\linewidth}}c@{\hspace{0.01\linewidth}}c@{\hspace{0.01\linewidth}}c@{\hspace{0.01\linewidth}}c@{\hspace{0.01\linewidth}}c@{\hspace{0.01\linewidth}}c@{\hspace{0.01\linewidth}}c@{\hspace{0.01\linewidth}}c@{\hspace{0.01\linewidth}}c@{\hspace{0.01\linewidth}}c}

\includegraphics[width = .11\linewidth]{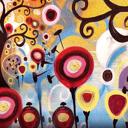} &
\hspace{1pt}\vrule\hspace{1pt}
\includegraphics[width = .11\linewidth]{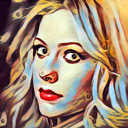} &
\includegraphics[width = .11\linewidth]{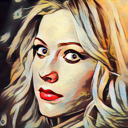} &
\includegraphics[width = .11\linewidth]{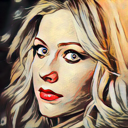} &
\includegraphics[width = .11\linewidth]{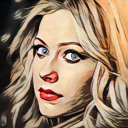} &
\includegraphics[width = .11\linewidth]{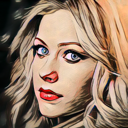} &
\includegraphics[width = .11\linewidth]{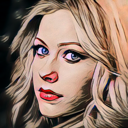} &
\hspace{1pt}\vrule\hspace{1pt}
\includegraphics[width = .11\linewidth]{figs/Transfer/style/11.jpg} & \\

\includegraphics[width = .11\linewidth]{figs/Transfer/style/12_1.jpg} &
\hspace{1pt}\vrule\hspace{1pt}
\includegraphics[width = .11\linewidth]{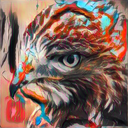} &
\includegraphics[width = .11\linewidth]{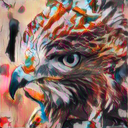} &
\includegraphics[width = .11\linewidth]{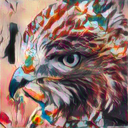} &
\includegraphics[width = .11\linewidth]{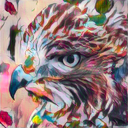} &
\includegraphics[width = .11\linewidth]{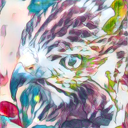} &
\includegraphics[width = .11\linewidth]{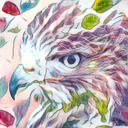} &
\hspace{1pt}\vrule\hspace{1pt}
\includegraphics[width = .11\linewidth]{figs/Transfer/style/04_1.jpg} & \\

\end{tabular}
}
\caption{Style interpolation (or transition) with the 16-style network. Images in the leftmost and rightmost are original styles. The content images used in the middle are from Figure~\ref{fig:Transfer16_results_selection_map1} and Figure~\ref{fig:Transfer16_diversity}.}
\label{fig:Transfer16_transition1}
\end{figure*}

%\vspace{-0.5em}
%\paragraph{Interpolation.}
%
With the selection unit, we interpolate between styles by adjusting the weights of different bits in the selection unit and generate the style interpolation (or transition) results in Figure~\ref{fig:Transfer16_transition1}. Specifically, if we denote $s_1$ and $s_2$ as the bit value of two styles and $N_1$ and $N_2$ as the corresponding noise map, the interpolation is generated by feeding $s_1\times N_1 + s_2\times N_2$ as the selection input.

\begin{figure}[t]
\centering
\small
{
\begin{tabular}{c@{\hspace{0.01\linewidth}}c@{\hspace{0.01\linewidth}}c@{\hspace{0.01\linewidth}}c@{\hspace{0.01\linewidth}}c@{\hspace{0.01\linewidth}}c@{\hspace{0.01\linewidth}}c@{\hspace{0.01\linewidth}}c@{\hspace{0.01\linewidth}}c@{\hspace{0.01\linewidth}}c@{\hspace{0.01\linewidth}}c}

\includegraphics[width = .175\linewidth]{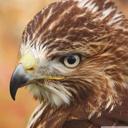} &
\hspace{1pt}\vrule\hspace{1pt}
\includegraphics[width = .175\linewidth]{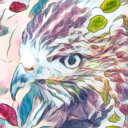} &
\includegraphics[width = .175\linewidth]{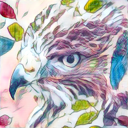} &
\includegraphics[width = .175\linewidth]{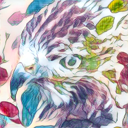} &
\hspace{1pt}\vrule\hspace{1pt}
\includegraphics[width = .175\linewidth]{figs/Transfer/style/04_1.jpg} &\\

\includegraphics[width = .175\linewidth]{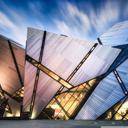} &
\hspace{1pt}\vrule\hspace{1pt}
\includegraphics[width = .175\linewidth]{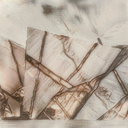} &
\includegraphics[width = .175\linewidth]{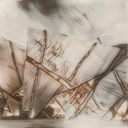} &
\includegraphics[width = .175\linewidth]{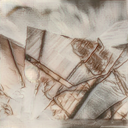} &
\hspace{1pt}\vrule\hspace{1pt}
\includegraphics[width = .175\linewidth]{figs/Transfer/style/14_1.jpg} &\\

%\hspace{1pt}\vrule\hspace{1pt}

%{(a)}& {(b)}& {(c)}\\
\end{tabular}
}
\caption{Diverse transferred results of our 16-style transfer network. Left: content images, Middle: diverse transferred results, Right: style images. Note the difference in the \emph{beak} and \emph{sky}.} %Results in the bottom row shows large but unreasonable variations with larger weight of the diversity loss.}
\label{fig:Transfer16_diversity}
\end{figure}

%\vspace{-0.5em}
%\paragraph{Diversity.}
%
Diverse transfer results are shown in Figure~\ref{fig:Transfer16_diversity}.
Different from the case of texture synthesis, the global structure of images is constrained by the demand of preserving content.
Therefore the diversity is exhibited at local visual structures.
Notice the slight but meaningful differences among these outputs.

\section{Discussion}

\begin{figure}[t]
\centering
\small
\begin{tabular}{c@{\hspace{0.01\linewidth}}c@{\hspace{0.01\linewidth}}c@{\hspace{0.01\linewidth}}c@{\hspace{0.01\linewidth}}c@{\hspace{0.01\linewidth}}c@{\hspace{0.01\linewidth}}c}

\includegraphics[width=4.1cm]{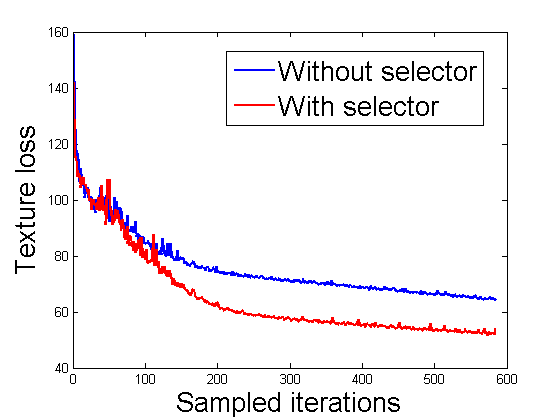} &
\includegraphics[width=4.1cm]{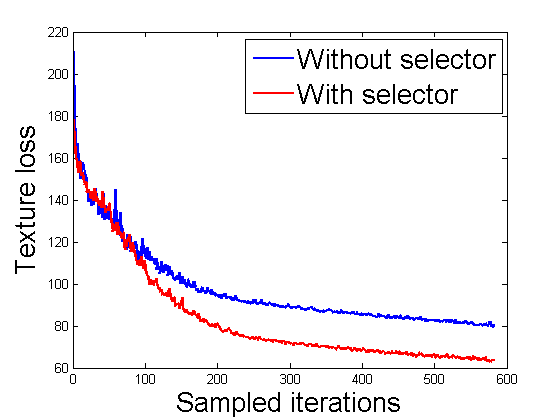} & \\

{ (a) Texture 20} & { (b) Texture 40}\\

\end{tabular}
\caption{Comparisons of the loss curve between our framework without and with the selector network.}
\label{fig:Selector}
\end{figure}

\paragraph{Selector network.}
In our model, we introduce a selector network (Figure~\ref{fig:framework}) in order to drive the network towards synthesizing the desired texture only.
%
%In experiments, we find that different bits in the selection unit are not enough to
%
The selector injects guidance to the generator at every upsampling scale and helps the model distinguish different textures better during the synthesis.
We show an example of training a 60-texture network w/o and w/ the selector network and compare these two settings in Figure~\ref{fig:Selector}. We present the loss curves of two textures as examples.
Figure~\ref{fig:Selector} clearly shows that with the selector, the network training achieves better convergence and thus generates better synthesized results.

%and thus generates better synthesized results.

\paragraph{Embedding.}
Starting with a one-hot selection unit to represent each texture with one bit, we first map it to a lower dimensional embedding via a linear projection and aim at learning a better representation of given textures.
In our presented 60-texture and 300-texture model, we map the 60-D and 300-D selection unit to a 32-D and 128-D embedding respectively.
Our synthesized results show that the embedding can still distinguish different textures for synthesis, which indicates that the original hand-crafted one hot representation is redundant.
In addition, as we have shown that new textures can be created through interpolation in a feed-forward way, it poses an open question that whether we can find the coefficients in a backward way to represent a given new texture as a weighted combination of learned embeddings.
%
%This is going to make a further stride towards the learning of arbitrary textures as well as images in general.
%
We leave this as a future direction to pursue.
%Our future work will focus on learning such embeddings for texture synthesis and style transfer.
%in understanding the computation behind style transfer networks as well as the representation of images in general.

\section{Conclusion}

In this paper, we focus on synthesizing multiple textures in one single network.
Given $M$ textures, we propose a deep generative feed-forward network which can synthesize diverse results for each texture.
%
%Users are provided with a selection unit to specify the desired texture for synthesis or create new textures by interpolation.
%
In order to train a deep network for multi-texture synthesis, we introduce the diversity loss and propose an incremental leaning scheme.
The diversity loss helps the network to synthesize diverse textures with the same input, and the incremental learning scheme helps effective and efficient training process.
Experimental results demonstrate the effectiveness of model, which generates comparable results compared to existing single-texture networks
but greatly reduces the model size.
We also show the extension of our multi-texture synthesis model to multi-style transfer for image stylization.

%\paragraph{\bf Acknowledgment.} This work is supported in part by the NSF CAREER Grant \#1149783, gifts from Adobe and Nvidia.

{\small
\bibliographystyle{ieee}
\bibliography{egbib}
}

\newpage

\end{document}